\documentclass[sigconf]{acmart}
\usepackage{amsfonts}
\usepackage{booktabs}
\usepackage{xspace}
\usepackage{multirow}
\usepackage{amsmath,bm}
\usepackage{tcolorbox}
\usepackage{graphicx}
\usepackage{booktabs}
\usepackage{color}

\usepackage{enumitem}
\usepackage{graphics}
\usepackage{epsfig}

\newcommand{\techname}{OCoR\xspace}

\usepackage{hyperref,amsmath,amssymb,amscd}

\AtBeginDocument{%
  \providecommand\BibTeX{{%
    \normalfont B\kern-0.5em{\scshape i\kern-0.25em b}\kern-0.8em\TeX}}}

\acmConference[]{ASE}{September 21--25, 2020}{Melbourne, Australia}
\acmPrice{15.00}
\acmISBN{978-1-4503-9999-9/18/06}

\setcopyright{acmcopyright}
\acmConference[ASE'20]{35th IEEE/ACM International Conference on Automated Software Engineering}{September 21--25, 2020}{Melbourne, Australia}
\acmBooktitle{35th IEEE/ACM International Conference on Automated Software Engineering (ASE'20), September 21--25, 2020, Melbourne, Australia}
\copyrightyear{2020}
\acmYear{2020}
\acmDOI{10.1145/1122445.1122456}

\begin{document}

\title{OCoR: An Overlapping-Aware Code Retriever}


\author{Qihao Zhu}
\affiliation{%
  \institution{Key Lab of High Confidence Software Technologies, Ministry of Education Department of Computer Science and Technology, EECS, Peking University}}
\email{Zhuqh@pku.edu.cn}

\author{Zeyu Sun}
\affiliation{%
  \institution{Key Lab of High Confidence Software Technologies, Ministry of Education Department of Computer Science and Technology, EECS, Peking University}}
\email{szy_@pku.edu.cn}

\author{Xiran Liang}
\affiliation{%
  \institution{Key Lab of High Confidence Software Technologies, Ministry of Education Department of Computer Science and Technology, EECS, Peking University}}
\email{liangxiran11@163.com}

\author{Yingfei Xiong}
\authornote{Corresponding author}
\affiliation{%
  \institution{Key Lab of High Confidence Software Technologies, Ministry of Education Department of Computer Science and Technology, EECS, Peking University}}
\email{xiongyf@pku.edu.cn}

\author{Lu Zhang}
\affiliation{%
  \institution{Key Lab of High Confidence Software Technologies, Ministry of Education Department of Computer Science and Technology, EECS, Peking University}}
\email{zhanglucs@pku.edu.cn}









\begin{abstract}
Code retrieval helps developers reuse code snippets in the open-source projects. Given a natural language description, code retrieval aims to search for the most relevant code relevant among a set of code snippets. Existing state-of-the-art approaches apply neural networks to code retrieval. However, these approaches still fail to capture an important feature: overlaps. The overlaps between different names used by different people indicate that two different names may be potentially related (e.g., ``message'' and ``msg''), and the overlaps between identifiers in code and words in natural language descriptions indicate that the code snippet and the description may potentially be related.

To address this problem, we propose a novel neural architecture named \techname\footnote{\techname, is short for An \underline{O}verlapping-Aware \underline{Co}de \underline{R}etriever.}, where we introduce two specifically-designed components to capture overlaps: the first embeds names by characters to 
capture the overlaps between names, and 
the second introduces a novel overlap matrix to represent the degrees of overlaps between each natural language word and each identifier. 

The evaluation was conducted on two established datasets. The experimental results show that \techname significantly outperforms the existing state-of-the-art approaches and achieves $13.1\%$ to $22.3\%$ improvements. Moreover, we also conducted several in-depth experiments to help understand the performance of the different components in \techname. 
\end{abstract}




\maketitle

\section{Introduction}
\label{intro}
Code retrieval is an important software engineering problem, which aims to retrieve the most related code snippet among a set of code snippets by a given natural language description. 
An effective code retriever helps developers reuse the code snippets from the internet. For example, if a SQL programmer gives an instruction ``get all the data in table A'', a code retriever will help the programmer to search from the large scale of code on the internet and find the target code ``select * from A''.

With the development of deep learning and the collection of large scale labeled datasets, neural networks have been widely used for various areas~\cite{vaswani2017attention,hu2015convolutional,gu2018deep,articlezhanglu,sun2019grammar}. For the task of retrieval, various approaches have been proposed~\cite{hu2015convolutional,Jian2016Convolutional,Hua2016Pairwise,gu2018deep,Yao2018Improving,Yao_2019},  by using neural networks. These approaches mostly embed the question and the answer into a high-dimensional vector space and try to find the most similar one between the vectors of questions and the vectors of answers (e.g., using cosine similarity). When it is applied to code retrieval, it takes the natural language description as the question and the target code as the answer~\cite{gu2018deep,Yao_2019}.   

\begin{figure}
    \centering
    \includegraphics[width =\linewidth]{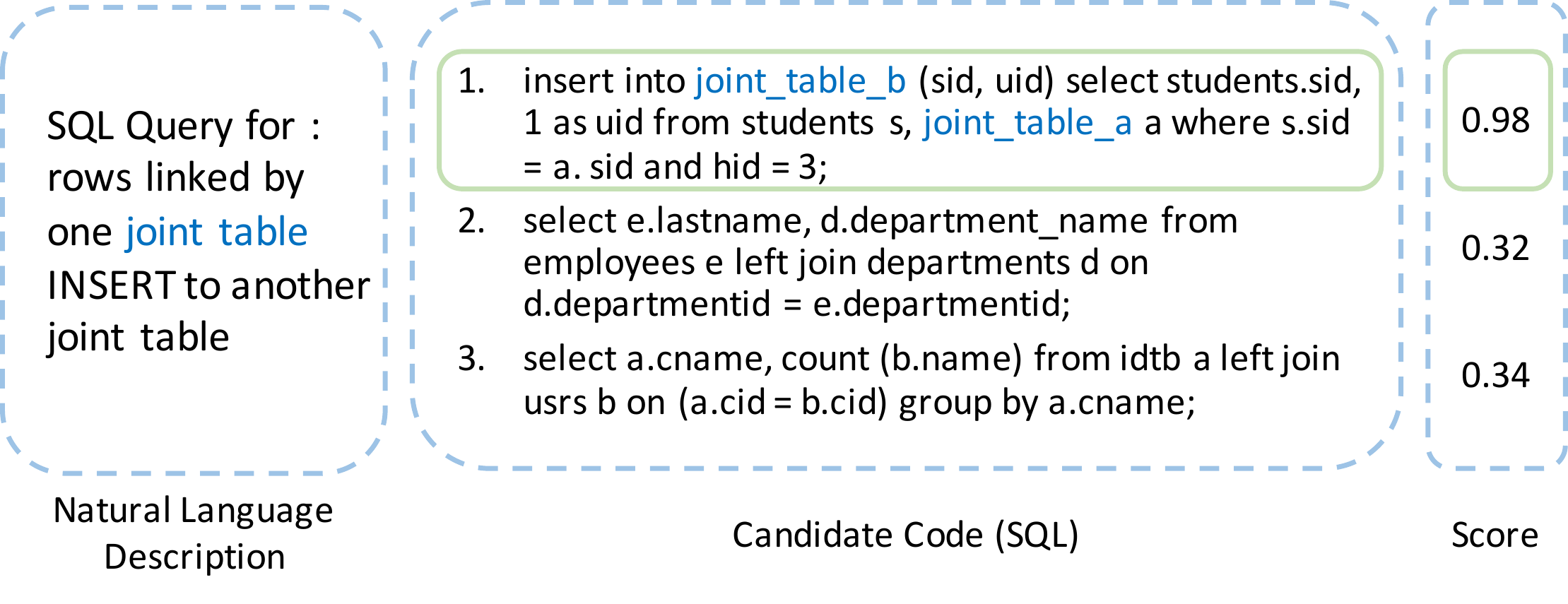}
    \caption{An Example from the StaQC dataset. The code retrieval in our approach ranks candidate code snippets with the scores given by the model.}
    \label{fig:example}
\end{figure}

However, these retrieval approaches fail to effectively handle overlaps, which are important in code retrieval. 
On one hand, different people may use different names to describe the similar meanings, either in code or in natural languages, and such names often have overlapped substrings. For example, ``Sort'' and ``QuickSort'' has the overlapped substring ``Sort''.
On the other hand, identifiers in code are often related to words in the natural language description. Though they may not be fully equal, overlapped sub-strings often exist. 
For example, in Figure~\ref{fig:example} the identifier ``joint\_table\_b'' is related to the words ``joint'' and ``table''. 
As far as we are aware, no existing neural architecture is specifically designed for handling overlaps.

To address these problems, we propose a novel neural architecture, \techname, a code retriever based on the overlap features. We represent each word by combining the representations of the characters within it, namely using the character-level embedding to capture the overlap between the names used by different programmers. Furthermore, we introduce a novel overlap matrix to represent the degrees of overlaps between each word in the natural language description and each identifier in code. Finally, we combine different code retrieval approaches by ensemble to enhance our model.

The experiment was conducted on several established datasets for SQL and C\# code retrieval, following \citeauthor{Yao_2019,iyer2016summarizing}~\cite{Yao_2019,iyer2016summarizing}. The experimental results show that our model significantly outperforms existing approaches by $13.1\%$ to $22.3\%$ improvements and achieve the best performance on all the datasets. To better understand our model, we also conducted the experiments focusing on the effectiveness of the components, and the results show that each component contributes to the overall performance. 

To summarize, this paper makes the following contributions:
\begin{itemize}
    \item We propose a novel neural architecture, \techname, for code retrieval. \techname uses two novel techniques, namely character-level embedding and the overlap matrix, to capture the overlaps between identifiers in code and words in natural language descriptions.
    \item We conducted extensive experiments to evaluate the effectiveness of our approach and the components in our approach. The results show that our approach significantly outperforms existing approaches by 13.1\% to 22.3\% improvements and all components in our approach are effective.
\end{itemize}
\section{Motivation}
\begin{figure}
    \centering
    \includegraphics[width =\linewidth]{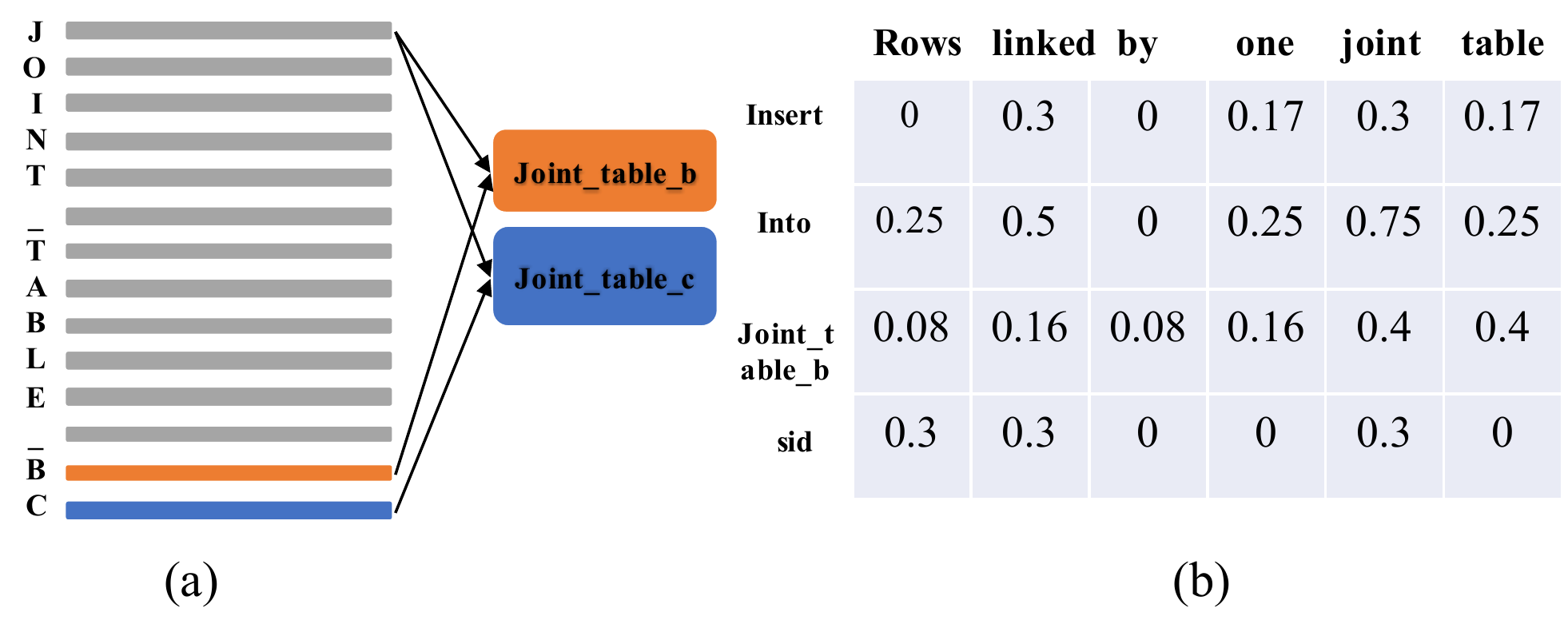}
    \caption{Examples of the computation of the character-level embedding and the overlap matrix.}
    \label{fig:motiv}
\end{figure}
As mentioned in the introduction, there are two types of overlaps in the code retrieval task. The first type of overlap is that different people may use different names to describe the similar meanings (e.g., the words in natural language and the identifiers in code). For example, ``joint\_table\_a'' and ``joint\_table\_b'' in the first SQL query in Figure~\ref{fig:example} could also be named as ``joint\_table\_1'' and ``joint\_table\_2''. 
If a neural network is trained over the code in Figure~\ref{fig:example}, it is difficult for it to know that the identifiers ``joint\_table\_1'' and ``joint\_table\_2'' are also related to the same query.
To address this challenge, existing approaches~\cite{Yao_2019} for code retrieval replacing variable names and raw strings with the variable types and numbers (e.g., rename the first table variable ``joint\_table\_b'' in a SQL with ``Table\_1''). 
In this way, the neural network is forced to ignore the identifier names but uses the structure of the code and the identifier types. 
However, the name of the identifier potentially carries useful information for the code retrieval, and ignoring them is likely to lower the performance.
To solve this problem, we propose character-level embedding to encode the names. The character-level embedding first encodes each character within each name via one-hot encoding and combine these relative vectors by a convolutional layer. Figure \ref{fig:motiv}(a) shows the computation of character-level embedding of ``joint\_table\_b'' and ``joint\_table\_c''. As shown, the combined vectors of these two identifiers are almost computed from the same vectors except the vector of the last character. Thus, the final embedding of these identifiers is closed to each other in the high-dimensional space.

The second type of overlap is that identifiers in code are often related to some words in natural language description. In a general perspective, this type of overlap is the overlap between question and answer, and is often considered in existing information retrieval approaches.
These approaches measure the number of the exactly matched tokens between the questions and the answers. However, in the code retrieval task, identifiers in code and words in natural language descriptions are often not fully equal. 
For example, as shown in Figure \ref{fig:example}, the identifier ``joint\_table\_b'' is not fully equal to any word in the natural language description, but it is related to two words ``joint'' and ``table''. 
To address this problem, we not only consider exactly matched words but also measure the degree of overlaps between partially matched words. 
We design a representation, named overlap matrix, to represent the degree of overlap between each word in natural language description and each identifier in code.
In this matrix, each row represents a word in the natural language description, while each column represents an identifier in code. 
Each cell is the degree of overlap between the word and the identifier. 
The degree of overlap can be measured by different metrics, and in this paper we use 
the longest common sub-string, the proportion of the longest consecutive sub-string $p$ that appears in both the natural languages word and the code identifier.
Figure \ref{fig:motiv}(b) shows a partial overlap matrix of the question-code pair in Figure \ref{fig:example}. 
We can see that though identifiers ``joint\_table\_b'' and word ``joint'' are not exact match, their degree of overlap is still higher than most other pairs.
Finally, our model takes the overlap matrix as input, utilizing the detailed overlap information for identifying the most related code snippet.

\section{Background}
\subsection{Convolutional Layer}
\begin{figure}
    \centering
    \includegraphics[width =\linewidth]{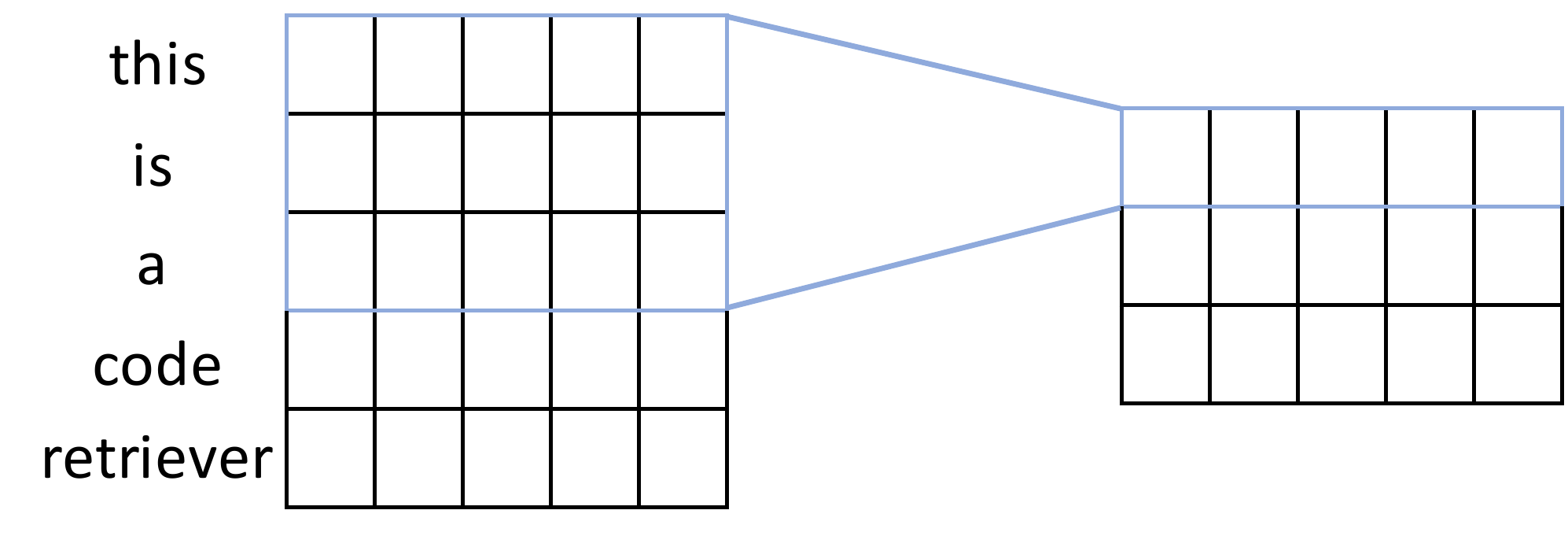}
    \caption{An Example of the computation convolutional layer.}
    \label{fig:conv}
\end{figure}
The Convolutional layer, which is the main building layer of a convolutional neural network (CNN)~\cite{krizhevsky2012imagenet}, has been widely used in various areas~\cite{krizhevsky2012imagenet,kim2014convolutional,hu2015convolutional,Jian2016Convolutional}. This layer can be regarded as a regularized version of a fully-connected layer. The fully-connected layer usually consists of several neurons, and each neuron in one layer is connected to all neurons in the next layer. Different from such a fully-connected layer, the connection in the convolutional layer is connected from each neuron in one layer to several corresponding neurons in the next layer. Such connections depend on pre-defined convolutional kernels.

In natural language processing, the convolutional layer is used to extract the features in the contents. Given an input vector, which represents the words in natural language, the convolutional layer uses kernels to extract the features in each vector and its neighboring vectors and outputs a new vector.  For example in Figure~\ref{fig:conv}, for an input vector of a sentence ``this is a code retriever'' with kernel size $3$, the layer outputs a new vector of ``is'' by a weighted summation of the vectors of ``is'' with its neighbors ``this'' and ``a''. This helps capture the features of the contents in the input natural language description and code of \techname. Thus, we apply it in our approach. The details of using this layer will be introduced in Section~\ref{sec:model}.

\subsection{Attention}
\begin{figure}
    \centering
    \includegraphics[]{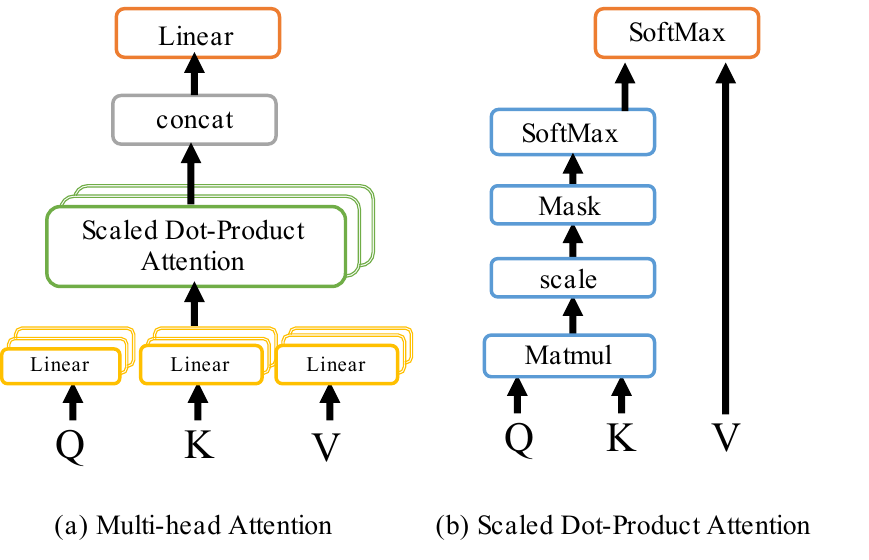}
    \caption{The detail of Multi-Head Attention.}
    \label{fig:att}
\end{figure}
In the basic encoder-decoder framework~\cite{sutskever2014sequence}\footnote{The encoder-decoder framework, which is also known as sequence-to-sequence framework, is a widely used approach in neural machine translation. In this framework, the neural network is divided into two parts: encoder, decoder. The encoder encodes the sequence into a vector and the decoder decodes the vector to a sequence in target language.}, the model always suffers from the the long-dependency problem~\cite{279181} with long sequences. To alleviate this problem, \citeauthor{bahdanau2014neural}~\cite{bahdanau2014neural} proposed the attention mechanism, which aims to let the neural model inspect the relevance between each pair of tokens in two long sequences.

Recently, to better alleviate the long-dependency problem, \citeauthor{vaswani2017attention}~\cite{vaswani2017attention} proposed a widely used attention mechanism called Multi-Head Attention. The overview of such mechanism is shown in Figure~\ref{fig:att}(a). This mechanism takes three vectors (mostly the representation of the words in the input sequence) as inputs and maps a query vector and a set of key-value vector pairs to the output. The main computation of this attention is called Dot-Product Attention layers, where the input of each layer consists of query vectors ($Q$), key vectors ($K$) and value vectors ($V$) as shown in Figure~\ref{fig:att}(b). The weights of the value vectors are calculated by the query vectors and the corresponding key vectors. Finally, the output is computed as a weighted sum of the values, where the query determines which values to focus on.

Self-attention is an attention mechanism for a single sequence to extract the complex comprehension of itself (formally, $Q = K = V$). This technique is often used to capture the long dependency information in sequences and has good performance in various tasks~\cite{vaswani2017attention,Xu2019LeveragingLA,Huang2019MusicTG}. The detailed computation of the attention mechanism will be introduced in the following section.


\section{Proposed Model}
\label{sec:model}
\subsection{Problem Definition}
We follow the existing studies~\cite{gu2018deep,Yao_2019} and use the same definition for code retrieval. Given natural language description $Q$ and a set of candidate code snippets $C$, our task is to retrieve a relevant code snippet $C^r \in C$ that specified by $Q$. 

As shown in Figure~\ref{fig:example}, to retrieve a code snippet, we first compute the relevance score between each code snippet $c \in C$ and the input natural language description $Q$. Then, we rank the code snippets in the set of candidate code snippets $C$. Finally, the code $C^r$ with the highest score is selected as the output of our approach, which is computed as
\begin{equation}
    C^r = \arg\max_{c \in C}R(Q, c)
\end{equation}
where $R$ denotes the computation of the relevance score. In our approach, the relevance score is a real number between 0 and 1. 

\subsection{Overview}
\begin{figure*}
    \centering
    \includegraphics[width =\textwidth]{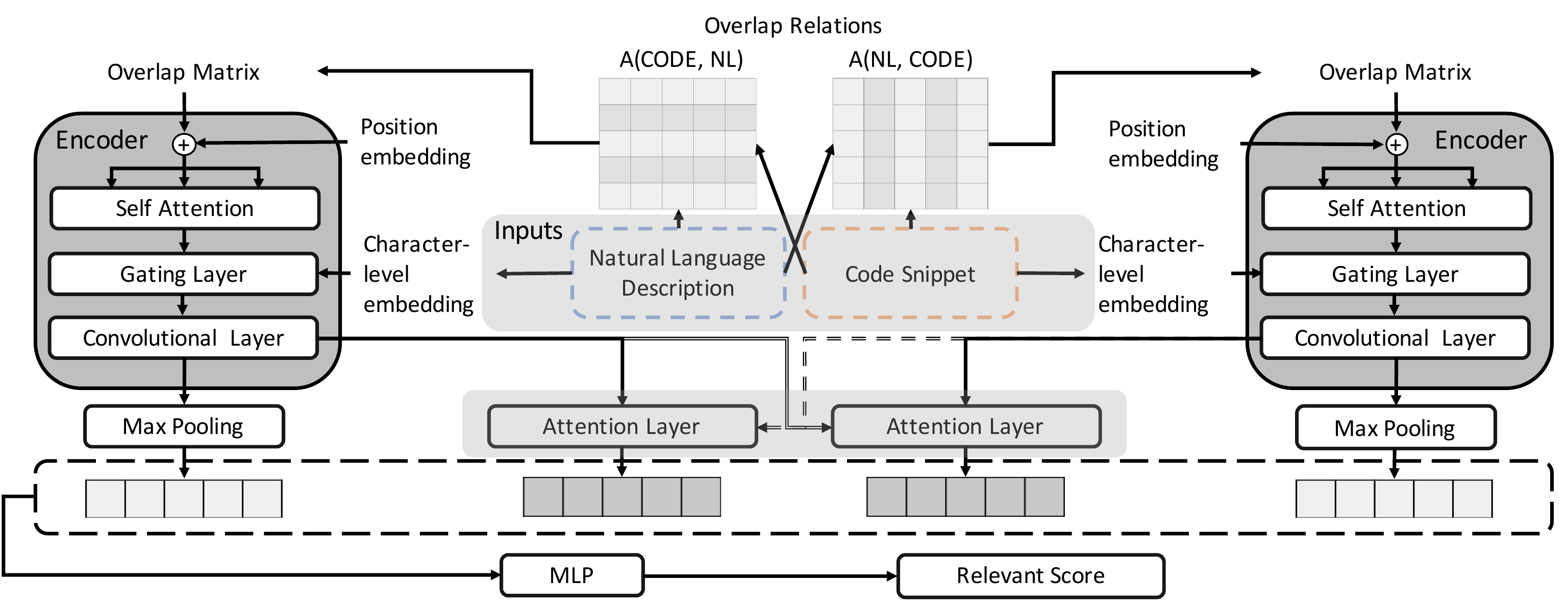}
    \caption{The overview of the \techname. The $A (CODE, NL)$ and $A (NL, CODE)$ are the Overlap Matrices.}
    \label{fig:overview}
\end{figure*}

Figure~\ref{fig:overview} shows the overview of \techname. We adopt the traditional overall architecture used in information retrieval \citet{Jian2016Convolutional}, where we encode the question (the natural language description) and the answer (the code snippet) respectively and combine the outputs via attention layers for further predicting the target relevant score. 
Based on this architecture, we design two encoders with the same structure for the question and the answer. Each encoder takes the overlap matrix and the question / answer as input and turns this input into a set of vectors.

Furthermore, as the evaluation will show later, 
our model complements existing approaches on code retrieval. To achieve even better performance, we use an an additional ensemble component to combine previous code retrieval models with \techname. 

In the rest of this section, we will describe the components of our architecture one by one. Besides,

\subsection{Input of Model}
The inputs of \techname are divided into three parts: 1) the natural language description; 2) the code snippet; 3) the overlaps between the natural language description and the candidate code snippet, where the last one is computed from the first two. 

\subsubsection{Preprocessing}
For the first two parts, we first process these inputs to make them suitable to be fed to the neural network. For the input natural language description, we first tokenize this input by the tool in the NLTK toolkit~\cite{Loper2002NLTK} and convert its characters within the tokens to lowercase. As for the input code snippet, we keep the original variable names and the raw strings, to keep the semantic information in the names. Then, we tokenize the code and also convert the characters within the names to lowercase. Thus, we get the preprocessed inputs for the neural network.



\subsubsection{Overlap Matrix}

As mentioned before, we use the overlap matrix to represent the degrees of overlaps between natural language words and code identifiers.
The overlap matrix is a real-valued matrix $A (T_1, T_2) \in \mathbb{R}^{L^{(T_1)} \times L^{(T_2)}}$, which contains the overlap scores between a token sequence $T_1$ and another token sequence $T_2$. Each cell $A_{ij}(T_1, T_2)$ in this matrix denotes the overlap score between the $i$-th token $T_1 (i)$ in the $T_1$ sequence and the $j$-th token $T_2 (j)$ in the $T_2$. Such score in \techname is computed by 
\begin{equation}
    A_{ij}(T_1, T_2) =
             \text{len}(S(T_1(i), T_2(j)) / \text{len}(T_2 (j))
    \label{eq:overlapscorf}
\end{equation}
where $\text{len}(T_2 (i))$ denotes the length of the word $T_1 (i)$, $S(T_1(i), T_2(j)$ denotes the longest common sub-string of $T_1, T_2$.
In particular, the computation of the overlap matrix $A$ is not commutative, i.e., $A(T_1, T_2) \not= A(T_2, T_1)$.
In our approach, we consider both the overlap score between the natural language description $NL$ and the code $CODE$ as well as the overlap score between the code and the natural language description, namely both $A(\bm{n}, \bm{c})$ and $A(\bm{c}, \bm{n})$ respectively. These two metrics are further fed to the encoder layer to extract features for code retrieval.


\subsection{Encoder}
In \techname, there are two encoders for both the natural language description and the code. Each encoder takes the overlap matrix and the natural language description / code as inputs. These inputs are encoded into vectors in a high-dimensional space for further similarity computations. 

To better encode the input information, inspired by \citeauthor{vaswani2017attention}~\cite{vaswani2017attention}, we design the encoder with a stack of $N$ mechanisms. Each mechanism contains three sub-layers: 1) a self attention layer; 2) a gating layer; 3) a convolutional layer. After each mechanism, the ResNet~\cite{he2015deep}\footnote{ResNet is residual learning framework to ease the training
of networks.} and the layer normalization~\cite{Ba2016Layer}\footnote{Layer normalization normalizes the values of the neurons into a suitable distribution, which eases the training.} are used. For the first mechanism, it takes the overlap matrix, namely $A(NL, CODE)$ / $A(CODE, NL)$, as input and further combines the natural language description / the code. For the rest of $N - 1$ mechanisms, they take the output of the previous mechanism as input and also combine the features of hidden layers in the natural language description / the code. We will first describe how we feed the overlap matrix to the first mechanism.

\paragraph{\textbf{Input Overlap Matrix.}}
The Overlap Matrix\footnote{We take $A(NL, CODE)$ as example in this section.} for our approach is a real-valued matrix, where each cell denotes the overlap scores between the natural language words and the identifiers in code. To take this matrix as input, we first reduce the matrix $A (NL, CODE)$ into an overlap vector $\bm{a}(NL)$, where the $i$-th element $\bm{a}_i(NL)$ in this vector denotes a new overlap score computed from the $i$-th row in the $A(NL, CODE)$ (the cells representing the $i$-th token in the natural language and each identifier in the code) via max-pooling. Max-pooling~\cite{kim2014convolutional} has been shown to be an effective way to reduce the matrix into a vector in various areas~\cite{giusti2013fast,zhou2016text,Yao_2019}. Following Max-pooling, we select the maximum value for each column as the element of $\bm{a}(NL)$, which is computed by 
\begin{equation}
    \bm{a}_i(\bm{n}) = \max_{j=1}^{\text{len}(\bm{c})}A_{ij}(\bm{n}, \bm{c})
\end{equation}

This vector contains the maximum overlap scores. For example, the fixed-size vector of the overlap matrix in Figure \ref{fig:motiv}(b) is $[0.3, 0.75, 0.4, 0.3]$.
To ease neural processing of the overlap scores, we further embed the scores with one-hot encoding. Please note that the scores are the real-values between 0 and 1. 
We partition the scores with the interval 0.01 and use a one-hot vector of length 100 to encode the scores.

\subsubsection{Self Attention Layer}
The first sub-layer in the encoder mechanism is the self attention layer. As we know, sequential information is important in both the natural language description and the code. To handle such information effectively, we apply the self attention mechanism, which is proposed by \citeauthor{vaswani2017attention}~\cite{vaswani2017attention} and has shown to be an effective way to encode this information~\cite{vaswani2017attention,Xu2019LeveragingLA}, as the first sub-layer in the encoder.

The self attention layer takes the previous output vectors $\bm{o}_1$,$\bm{o}_2$,$\cdots$,$\bm{o}_{L}$ as input, where $L$ denotes the length of the input natural language description / code. This layer consists of two parts: 1) the position embedding layer; 2) multi-head attention layer.
\paragraph{\textbf{Position Embedding Layer.}} A position embedding layer is a standard layer in transformer architecture~\cite{vaswani2017attention} to provide the indexes of the words in the input matrix. 
For example, the layer should know the word ``joint'' is the $5$-th token in the natural language description in Figure \ref{fig:example}.
If we directly use an attention layer for the input vectors, the position information will not be considered, and this is why we need the position embedding layer. 

In this layer, the vector of the $i$-th position is represented as a real-valued vector, which is computed by
\begin{equation}
        \begin{aligned}
                 p_{(i,2j)} &= 
                    \sin (pos/(10000^{2j/d} )) \\ 
                     p_{(i,2j + 1)} & = \cos (pos/(10000^{2j/d} )) 
        \end{aligned}
\end{equation}
where $pos = i + step$, $j$ denotes the element of the input vector and $step$ denotes the embedding size. After we get the vector of each position, we directly add this vector to the corresponding input vector, where $\bm{e}_i = \bm{o}_i + \bm{p}_i$.

\paragraph{\textbf{Multi-head Attention Layer.} }
The second part of the self attention layer is the multi-head\footnote{The multi-head mechanism consists of several heads, each of which is an attention layer, separately. The outputs of these heads are further jointed together by a fully-connected layer.} attention layer. As introduced in the background section, an attention mechanism maps a query, a key and a value to an output. In this layer, the query, the key, the value, and output are all vectors. 

Following the definition of \citeauthor{vaswani2017attention}~\cite{vaswani2017attention}, we divide the multi-head mechanism into $H$ heads. Each head is an attention layer, which maps the query $Q$, the key $K$ and the value $V$ to an output, namely the output of each head $head$. The computation of the $s$-th head is represented as
\begin{equation}
    head_s = \text{softmax}(\frac{QK^T}{\sqrt{d_k}})V
\end{equation}
where $d_k = d/H$ denotes the length of each extracted feature vector and $Q$, $K$ and $V$ are computed by a fully-connected layer from $Q$, $K$, $V$. In the encoder, the vectors $Q$, $K$ and $V$ are all the outputs of the position embedding layer $\bm{e}_1, \bm{e}_2, \cdots, \bm{e}_L$. The outputs of these heads are further jointed together with a fully-connected layer, which is computed by
\begin{equation}
    Att = [head_1;\cdots;head_H]\cdot W_h
\end{equation}
where $W_h$ denotes the weights in fully-connected layer and the output vectors $Att = [\bm{a}_1, \bm{a}_2, \cdots, \bm{a}_L]$ are the high-level vectors, which combine the sequential information and the original information together. However, these vectors still fail to encode the semantic information of each word effectively at least in the first mechanism in the encoder. Thus, we will then describe how we address this issue via a gating layer .

\subsubsection{Gating Layer}
The second sub-layer in the encoder mechanism is a gating layer. This layer takes the outputs of the previous layer and the input natural language description / code as input. existing state-of-the-art approaches~\cite{Yao_2019} use the word2vec~\cite{mikolov2013efficient} mechanism to utilize the semantic information of the input. However, it may be not suitable for code retrieval, where the similar identifiers can be named differently by different programmers but with overlapped characters (e.g., it may have a very similar meaning for two words ``dataId'' and ``data\_id''), which may lead to a large number of vocabulary for neural networks to learn. To address this issue, we propose to use the character-level semantics to catch the overlaps between identifiers in code retrieval. We combine the outputs of the previous layer with the character-level embedding approach for each word. 

\paragraph{\textbf{Character Embedding}} To implement the character embedding, we first pad each token (both the word in the natural language description and the identifier in code) to a fixed length $CL$ with a special character. In particular, if the length of the token is more than $CL$, we truncate the end of this token and make it a $CL$-length token. Then, we represent each character in the token as a real-value vector, namely $embedding$. As we know, a token consists of several characters. To catch the semantic information of each token, we adopt a set of convolutional layers to integrate the vectors of the characters within the token. The extracted semantic vector for the $i$-th token $\bm{t}_i$ is computed by
\begin{equation}
    \bm{t}_{(i, n)} = W_{(c, n)}[\bm{t}_{(1, n - 1)}; \bm{t}_{(2, n - 1)};\cdots;t_{(CL, n - 1)}]
\end{equation}
where $W_c$ are the convolutional weights and $n$ denotes the $n$-th layer of the convolutional layers. In particular, $\bm{t}_{(k, 0)} = \bm{c}_k$, where $\bm{c}_k$ denotes the character embedding vector of the $k$-th character within the $i$-th token. In our approach, we have three convolutional layers for this character embedding layer. For the first two convolutional layers, we use the zero padding and the sizes of the convolutional kernel are set to 3 and 5, respectively.
\paragraph{\textbf{Gating Mechanism}} To incorporate the semantic information of each token with the previous outputs, we use a mechanism named Gating Mechanism~\cite{DBLP:journals/corr/abs-1911-09983}. This mechanism incorporates an input semantic vector $\bm{t}_i$ with a given control vector\footnote{The control vector is the special vectors given in our approach. This vector decides the weights of different vectors.} with the multi-head mechanism. In this paper, we use the previous output vectors, namely $\bm{a}_i$, as the control vector.
The computation of the gating layer in our model can be represented as
\begin{equation}
    \alpha_i^o = exp(\bm{q}_i^T\bm{k}_i^o)/\sqrt{d}
\end{equation}
\begin{equation}
    \alpha_i^c = exp(\bm{q}_i^T\bm{k}_i^c)/\sqrt{d}
\end{equation}
\begin{equation}
    \bm{h}_i = (\alpha_i^o\cdot\bm{v}_i^o + \alpha_i^c\cdot\bm{v}_i^c)/(\alpha_i^o + \alpha_i^c)
\end{equation}
\begin{equation}
    head_s = [\bm{h}_i;\cdots;\bm{h}_H]
\end{equation}
where $\bm{q}_i$, $\bm{k}_i^o$, $\bm{v}_i^o$ are all computed by a fully-connected layer over the control vector $\bm{a}_i$; $\bm{k}_i^c, \bm{v}_i^c$ are computed by another fully-connected layer over the semantic vector $\bm{t}_i$. After this computation, we enhance the vectors with the semantic information, and the extracted new features are denoted as $\bm{c}^{(com)}_1, \bm{c}^{(com)}_2, \cdots, \bm{c}^{(com)}_L$.
\subsubsection{Convolutional Layer}
The final sub-layer in the encoder mechanism is a set of convolutional layers. We follow the design of the encoder proposed by \citeauthor{vaswani2017attention}~\cite{vaswani2017attention} and adopt a set of convolutional layers to extract the local features around each token.
The computation of the convolution layer can be represented as
\begin{equation}
                    \bm{y}^{l}_i = W_{l}[\bm{y}^{l - 1}_{i - w};\cdots;\bm{y}^{l - 1}_{i + w}]
                    \label{eq:conv}
\end{equation}
where $l$ denotes the $l$-th convolutional layer in the set, $W_l$ are the convolutional weights, $w = (k - 1) / 2$ and $k$ denotes the window size. 
In particular, $\bm{y}^{l - 1}_{i}$ is the output of the previous gating layer $\bm{c}^{(com)}_i$. We use two convolutional layers in this sub-layer and add the activation function $GELU$~\cite{hendrycks2016bridging} between these convolutional sub-layers. In particular, we use the zero padding in these layers.

\subsection{Max Pooling}
After all these $N$ mechanisms in the encoder, we adopt an additional convolutional layer as Equation~\ref{eq:conv}, which is padded with a special vector during convolution. Then, we get the final features of each token. The features denote the high-level information of the input natural language description / code. However, these features have the same shape as the input. To facilitate further prediction for code retrieval, we need to aggregate such features into a fixed-size vector, which is regardless of the input size.

Max pooling has shown the power in aggregating features, thus, we apply the max pooling approach over the outputs of the encoder and extract the fixed-size vector for each encoder.



\subsection{Attention Layer}
The encoders encode the information of the input natural language description and the input code separately. However, it still lacks the relations between two inputs even if we have used the prior knowledge of the overlaps. To help the neural network learn such relations between the two inputs, we adopt two attention layers after the encoder. 

As described in the previous section, the outputs of the encoder combine the overlap information with semantic information (character embedding). Thus, we also apply the multi-head attention layer to the outputs of two encoders to extract the relations. As for the description and the code, we design two separate attention layers for them. The computation  of the attention mechanism is similar to the ``self attention'' with different inputs. One layer treats the encoding of the description as query ($Q$) and the other treats the encoding of the code as query ($Q$). This design allows the model to extract the weighted sum of the outputs of two encoders based on each other. After the attention, two convolutional layers and a max pooling layer are followed to integrate the features. 

\subsection{Prediction}
After all the computation of the attention layer, we concatenate all features. They are further fed to a two-layer perceptron followed by a {\it softmax} activation. The output of these computation is the classification probability of two classes. The first class denotes that the input natural language description and the input code is related, whereas the second class denotes that the input natural language description and the input code is not related.  In our approach, the predicted classification probability of the first class is the relevance score between the input natural language description and the code, where the relevance score is computed by

\begin{equation}
R(Q, c)= \frac{\exp\{h_1\}}{\sum_{j=1}^2 \exp\{h_j\}}
\end{equation}
where $h_i$ is the input logit of {\it softmax}.   


\subsection{Training}
Our model is trained by minimizing cross-entropy loss against the ground truth. Specifically, for each training data $<Q, C, A>$ where $Q$ is the description, $C$ is the code, $A$ denotes the ground truth class. The cross-entropy loss is computed by
\begin{equation} 
    Loss(\theta) = - \sum_{i=1}^2 g(i)*\log{\theta(i)}
\end{equation}
where $g$ denotes the ground truth class, $\theta$ is the classification result predicted by the neural network.

\subsection{Model Combination}
On the basis of our basic model introduced above, we also consider an additional method that combines different models on code retrieval task together by ensemble. Inspired by \citeauthor{Yao_2019}~\cite{Yao_2019}, where the proposed CoaCor combines code retrieval with code annotation for better performance. We consider to combine different models by integrate the relevance scores computed by different models and output the final relevance score for \techname. The score is computed by a linear combination as
\begin{equation}
     R(Q, c) = \lambda * S_1 + (1 - \lambda) * S_2
\end{equation}
where $S_1$ denotes the relevance score computed by \techname, $S_2$ is the score computed by the combined model, $\lambda$ is a real number between 0 and 1.

\section{Experiment Setup}
In this section, we present the experimented setup. We will first introduce the research questions\footnote{The code of our experiment is available at https://github.com/anyone546/OCoR.}.  

\subsection{Research Question}
Our evaluation aims to answer the following research questions:
\begin{itemize}
    \item \textbf{RQ1 What is the performance of \techname?}\\
    To answer this question, we conducted an experiment on several established datasets and compared the performance of \techname with the existing state-of-the-art approaches.    
    \item \textbf{RQ2 What is the contribution of each component in \techname?}\\
    To answer RQ2, we start from the full model of \techname, and 
    in turn removed each component to understand the contribution of it. Then, we also replaced the metrics used in measuring the overlap score with longest common prefix (LCP) and word embedding based similarity to better understand the contribution of the component of the overlap matrix.
    \item \textbf{RQ3 Why does the model combination work?}\\
    In fact, the result of RQ1 will suggest that the model combination places a significant role in the overall performance. To understand why different models can be combined, we analyzed the distribution of the result predicted by different models on the SQL dataset. More specifically, we selected some examples from these datasets to show the differences of the models. 
\end{itemize}
\subsection{Dataset}
\begin{table*}
  \begin{tabular}{lcccccccc}
    \toprule
    Statistics&StaQC-train&StaQC-val&StaQC-test&DEV&EVAL&C\#-train&C\#-dev&C\#-test\\
    \midrule
    Number of QC-pairs & 89,688 & 11,932 & 17,899 & 330 & 300 & 77,816&17,849&17,849\\
    Number of Cases & - & 11,900 & 17,850 & 6,600 & 6,000 & - &17,800&17,800\\
    \midrule
    Avg. tokens in description & 9 & 9 & 9 & 10 & 15&12&12&12\\
    Max. tokens in description & 32 & 35 & 45 & 45 & 35&&37&34\\
    Avg. tokens in code & 59 & 62 & 60 & 47 & 47&38&38&38\\
    Max. tokens in code & 3,367 & 2,774 & 2,672 & 291 & 291&290&300&310\\
  \bottomrule  
\end{tabular}
\caption{Statistics of the datasets we used. ``Number of QC-pairs'' denotes the total amount of question-code pairs in the specific dataset. ``Number of Cases'' is the number of the retrieve cases for evaluation. 
}
\label{tab:sta}
\end{table*}

Our experiment is based on two established benchmarks: the StaQC benchmark\cite{Yao_2018}, and the C\# benchmark used in \citeauthor{iyer2016summarizing}~\cite{iyer2016summarizing}. The StaQC benchmark contains 119,519 question-code pairs written in the SQL. These pairs are collected from Stack Overflow~\cite{stackoverflow}, making itself the largest-to-date in SQL domain. 
We followed the original train-dev-test split in StaQC, namely StaQC-train, StaQC-val ,and StaQC-test. For better evaluation , we used two additional test dataset namely ``DEV'' and ``EVAL'' for SQL, are collected by \citeauthor{iyer2016summarizing}~\cite{iyer2016summarizing}. These datasets contain 110 and 100 code written in SQL respectively. For every snippet, they use three different references written by humans as the additional test cases. 
The second benchmark contains 113,514 question-code pairs written in C\# collected from StackOverflow. We split the dataset into C\#-train, C\#-val ,and C\#-test as \citeauthor{iyer2016summarizing}~\cite{iyer2016summarizing,Yao_2019}
The detailed statistics of these datasets are listed in Table~\ref{tab:sta}. 

For the StaQC benchmark, we took the training set of StaQC-train as the training set, took the DEV set as the development set, and took other three datasets, StaQC-test, StaQC-val, ``EVAL'', as the test set. For the C\# benchmark, we also followed the same experiment settings during training and the C\#-val was treated as the development set in our experiment. 

Note that to test the performance of a code retrieval approach, we not only need the desirable question-code pair (the positive answer), but also other code snippets as negative answers. We call such a case containing a question and a set of code snippets as a \emph{retrieval case}. We used the same retrieval cases used in existing works~\cite{iyer2016summarizing,Yao_2019}, where the counts of these cases are show in the ``Number of Cases'' row in Table~\ref{tab:sta}. Each retrieval case contains 1 positive code snippet and 49 negative code snippets.


\subsection{Metrics}
To measure the performance of our approach, we followed  \citeauthor{Yao_2019}~\cite{Yao_2019} and used a standard metrics called Mean Reciprocal Rank (MRR)~\cite{craswell2009mean} in this paper. 


The MRR metrics is computed over the entire dataset \\
$D = \{(Q_1, C_1), (Q_2, C_2), \cdots, (Q_n, C_n)\}$:
\begin{equation}
    MRR = \frac{\sum_{i=1}^n 1/r_i}{|D|}
\end{equation}
where $r_i$ denotes the ranking of $C_i$ in the $i$-th query $Q_i$. In this metrics, the higher value denotes the better performance of code retrieval.

\subsection{Implementation Details}
We implemented our approach based on Tensorflow~\cite{Abadi2016TensorFlow}. We set the $N = 3$, which denotes that each encoder in our experiment contains a stack of 3 mechanisms. We set the embedding size for both the characters and the overlap scores to $256$. All hidden sizes were set to $256$ except that $1024$ was used for both the first layer of the convolutional layer and the first layer in the MLP. During training, the dropout~\cite{hinton2012improving} was used to avoid overfitting, where the droprate was set to $0.2$. Our model was optimized by Adam optimizer~\cite{kingma2014adam} with learning rate $0.0001$. For the model combination, we set the hyper-parameter $\lambda$ to $0.1$. These hyper-parameters and parameters for our model were chosen based on the development set (DEV is used), which followed the existing state-of-the-art work~\cite{Yao_2019}. Specifically, for each query natural language description in the training corpus, we randomly sampled 5 code snippets as the negative examples for each training epoch. In \techname, the natural language description and the code snippet shared the same embedding weights.
\subsection{Baselines}
In our experiment, we used the existing state-of-art code retrieval approaches as the baselines for comparison.
\begin{itemize}
    \item Deep Code Search (DCS)~\cite{gu2018deep}. DCS jointly embeds the input code snippets and the input natural language description into a high-dimensional vector space with an RNN based neural network. In this way, a code snippet and its corresponding natural language description have similar vectors, which are then used for computing the similarity between two inputs by cosine similarity.
    \item CODE-NN~\cite{iyer2016summarizing}. The core component of CODE-NN is an LSTM-based RNN~\cite{hochreiter1997long} with attention. This attention mechanism computes the probability of an natural language description, given a code snippet. For code retrieval, given an input natural language description, CODE-NN computes the probability of the input for each code. After the computation, CODE-NN ranks the given code snippets based on the probability. 
    
    \item CoaCor~\cite{Yao_2019}. CoaCor used a reinforcement learning-based framework to combine code retrieval and code annotation together for enhancing the code retrieval. They also combine the code retrieval approaches together by ensemble to improve the performance.
    In particular, the basic model of CoaCor is denoted as QN-RL$^{\text{MRR}}$, whereas the combined models (the best performance) are denoted as QN-RL$^{\text{MRR}}$ + CODE-NN.
    
\end{itemize}
\section{Results}
\begin{table*}
  \begin{tabular}{l|lcccc}
    \toprule
    & Model&EVAL&StaQC-val&StaQC-test&C\#\\
    \midrule
    \multirow{4}{*}{\rotatebox{90}{Ori.}}			
    & DCS & 0.555&0.534&0.529&0.441\\
    & CODE-NN  & 0.514 & 0.526 & 0.522&0.531\\
    & QN-RL$^{\text{MRR}}$  & 0.512 & 0.516 & 0.523 &0.528\\
	\cmidrule(rrrrr){2-6}
    & \techname  & \textbf{0.601} & \textbf{0.647} & \textbf{0.643}& \textbf{0.682}\\
    \midrule
    \midrule
    \multirow{3}{*}{\rotatebox{90}{Com.}}		
    & QN-RL$^{\text{MRR}}$ + CODE-NN  & 0.571 & 0.575 & 0.576&0.629\\
	\cmidrule(rrrrr){2-6}
    & \techname + QN-RL${^\text{MRR}}$  & 0.630 & 0.658 & 0.677&0.746\\
    & \techname + (QN-RL${^\text{MRR}}$ + CODE-NN) & \textbf{0.646} & \textbf{0.665} & \textbf{0.685}&\textbf{0.764}\\
  \bottomrule
    \end{tabular}
  \caption{The results show the MRR of code retrieval among 50 examples. In this table, we divide the existing model into two different categories. The first category (Ori.), where the original model is used, is in the 2 to 5 rows. The second category (Com.), where different models are combined together by ensemble, is in the 6 to 9 rows.}
  \label{tab:freq}
\end{table*}
In this section, we report the results of our experiment and answer the research questions.
\subsection{Performance of \techname (RQ1)}
The results of RQ1 are presented in Table~\ref{tab:freq}.

We first compare the original \techname with the original models (Ori. in Table~\ref{tab:freq}). As shown, among the three state-of-the-art models, \techname achieves the best performance on all datasets. \techname is higher than the existing best results, by $9.1\%$ to $28.2\%$ improvements. 

For the model combination, we first combine \techname with the original model QN-RL$^\text{MRR}$ (denoted as OCoR + QN-RL$^\text{MRR}$). We select the state-of-the-art models proposed by \citeauthor{Yao_2019}~\cite{Yao_2019} (QN-RL$^{\text{MRR}}$ + CODE-NN) as the baselines. As shown in Table~\ref{tab:freq}, our approach significantly outperforms existing state-of-the-art models by 10 points improvement on average. In particular, we achieved $13.1\%$ to $22.3\%$ improvements on all datasets, which shows effectiveness of our approach. Then, we combine \techname with the model combined by \citeauthor{Yao_2019} (QN-RL$^{\text{MRR}}$ + CODE-NN), where we also achieved the best results (OCoR + (QN-RL$^\text{MRR}$ + CODE-NN)) among all datasets. 

The results suggests that our approach is more effective than existing state-of-the-art models on different datasets and different program languages.

\begin{tcolorbox}
Answer to \textbf{RQ1}: \techname has a good performance ($13.1\%$ to $22.3\%$ improvements) compared with the existing state-of-the-art approaches on all datasets covering two programming languages.
\end{tcolorbox}


\subsection{The contribution of each component (RQ2)}
\begin{table}
\resizebox{8.5cm}{!}{
  \begin{tabular}{lccc}
    \toprule
    Model&EVAL&StaQC-val&StaQC-test\\
    \midrule
    \techname  & \textbf{0.601} & \textbf{0.647} & \textbf{0.643}\\
    - Overlap Score  & 0.420 & 0.545 & 0.538\\
    Character-level Overlap $\xrightarrow{}$ WordSimilarity   &0.554& 0.603&0.605 \\
    \midrule
    \ Overlap $\xrightarrow{}$ LCP & 0.591 &0.628&0.632\\
    
    \bottomrule
\end{tabular}}
  \caption{The ablation test on \techname, where \techname denotes the original model of our approach.}
    \label{tab:abla}

\end{table}
To answer RQ2, we first conducted the ablation test on the SQL dataset to figure out the contribution of each component. In this subsection, we only conducted the experiment based on the original \techname, which aims to understand the contribution of each component more clearly.

The model in the ablation test had the same setting with the original \techname except we removed each component in turn. The results are presented in Table~\ref{tab:abla}. We first removed the input overlap scores from \techname. To remove this, instead of using overlap matrices as the inputs of the first mechanism in the encoder, we replaced it with the input tokens in natural language description / code. Furthermore, we followed the previous joint model ~\cite{gu2018deep,cambronero2019deep} and the model was trained by minimizing the cosine similarity between the natural language description and code 
as the . By applying such settings, the performance was closed to the previous approach based on word2vec which shows the effect of the overlap matrix. 


 To better understand the contribution of the metrics for computing overlap scores (namely Equation~\ref{eq:overlapscorf}), we further conducted the experiment on the same datasets but with a different metrics of overlap scores.
The baseline metric compared with the original metric is word similarity based on word embedding. We first used the GloVe~\cite{pennington2014glove} to pretrain the word embedding vector based on the training set with BPE~\cite{sennrich2015neural}. Then we use the cosine similarity of $\bm{w}_1, \bm{w}_2$ as the overlap score. The results are presented in Table~\ref{tab:abla}. The character-level metric achieves better performance on all datasets. This result shows our metrics are more suitable than the traditional similarity matrix used in document retrieval for code retrieval.

We also replaced the original overlap metric with Longest Common Prefix (LCP). The longest common prefix of a pair of strings $a$ and $b$ is the longest string $p$ which is the prefix of both strings. We use $len(p) / len(a)$, $len(p)/ len(b)$ as the overlap scores of two strings, respectively. In particular, we also compute the score based on the longest common suffix, and finally select the bigger one as the overlap score. The performance of this metric is slightly lower. The ablation test also indicates that the character-level information is important to the code retrieval task.
\begin{tcolorbox}
Answer to \textbf{RQ2}: Each component contributes to the overall performance of \techname.
\end{tcolorbox}



\subsection{Model Combination Analysis (RQ3)}

\begin{table*}
  \begin{tabular}{lp{14cm}}
    \toprule
    Type&Example\\
    \midrule
    \midrule
    Description &\ {\color{magenta}\textbf{select}} a formatted {\color{orange}\textbf{date}} range from values in a table column\\
    \midrule
    SQL Code &\ {\color{magenta}\textbf{select}}	{\color{orange}\textbf{date}}\_format	(start{\color{orange}\textbf{date}},	'\%m')	+	{\color{orange}\textbf{date}}\_format	(start{\color{orange}\textbf{date}},	'\%d')	+	'-' +	{\color{orange}\textbf{date}}\_format	(end{\color{orange}\textbf{date}},	'\%d')	+	','	+	{\color{orange}\textbf{date}}\_format	(start{\color{orange}\textbf{date}},	'\%y')	from	yourtable;\\
    \midrule
    \midrule
    Description &\ SQL {\color{magenta}\textbf{Insert}} multiple Values {\color{orange}\textbf{where}} 1 value comes {\color{blue}{\textbf{from}}} a {\color{orange}\textbf{select}} query\\
    \midrule
    SQL Code &\ {\color{magenta}\textbf{insert}}  into  table2  (  telnumber  ,  adress  )  {\color{orange}\textbf{select}} '12324567890'  ,  applicatieid  {\color{blue}{\textbf{from}}}  applicatie  {\color{orange}\textbf{where}}  name  =  'piet'  ;\\
    \midrule
    \midrule
    Description &\ Quick way to {\color{magenta}\textbf{space}} fill column {\color{orange}\textbf{256}} chars SQL-Server 2012\\
    \midrule
    SQL Code &\ select  {\color{magenta}\textbf{space}}  (  {\color{orange}\textbf{256}}  )  ;\\
    \bottomrule
\end{tabular}
  \caption{The examples that ranked perfectly by \techname but not CoaCor.}
    \label{tab:examples_ocor}
\end{table*}
\begin{figure}
    \centering
    \includegraphics[width =\linewidth]{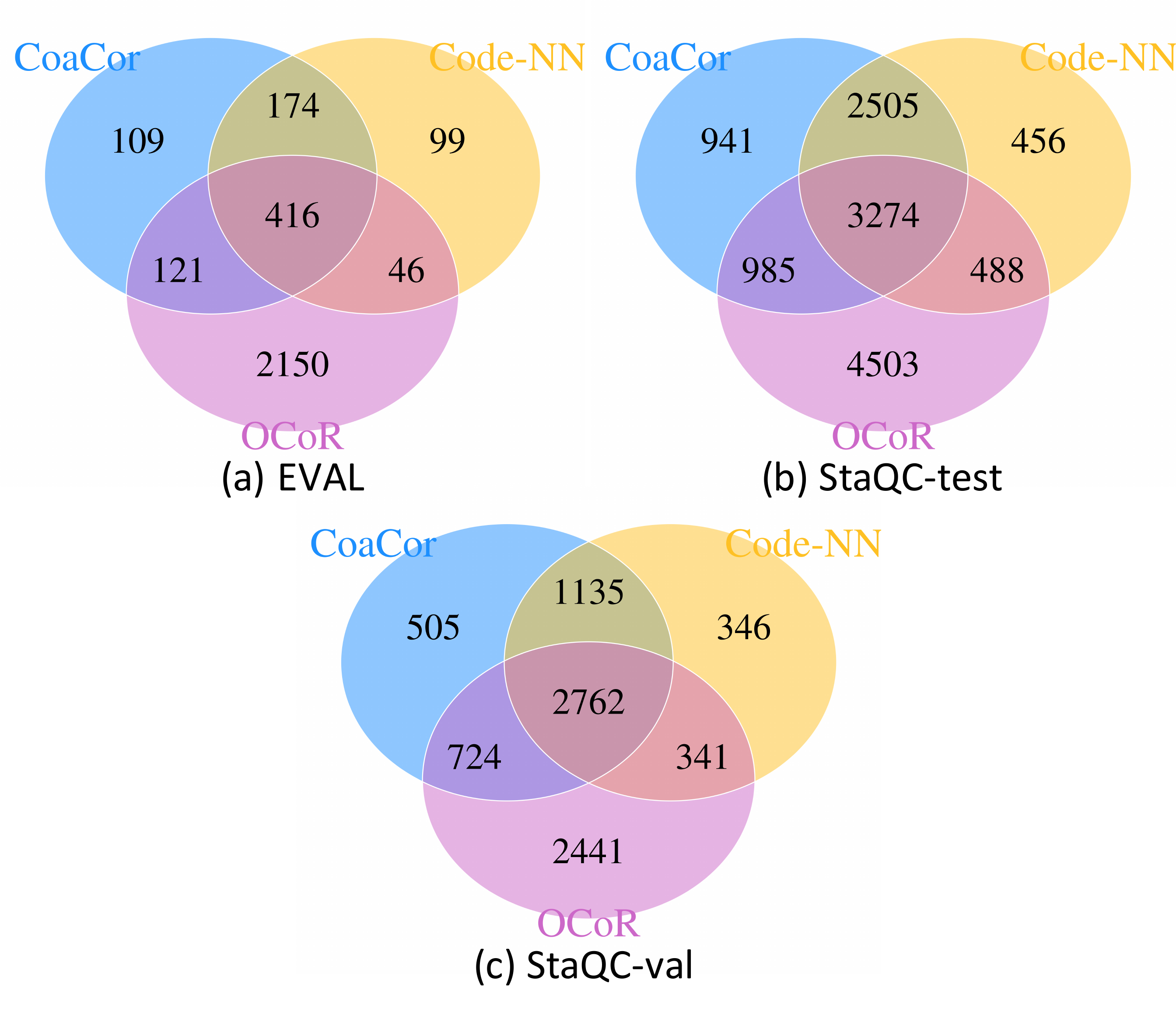}
    \caption{The overlaps of the prefect ranking of different approaches the among three datasets.}
    \label{fig:overlap}
\end{figure}
To answer RQ3, we try to figure out the reason why the model combination works. We first implemented the existing approaches, CoaCor (QN-RL$^\text{MRR}$) and CODE-NN. Then, we 
studied the prediction overlaps of the perfect ranking among the datasets. For a given natural language description and a set of code snippets with one positive code snippet, the ``perfect ranking'' in this paper means that the positive code snippet is ranked in the top-1.

The results are presented in Figure~\ref{fig:overlap}. As shown, on these three datasets, $16.2\%$ perfect ranking cases on average can be solved by all three approaches, whereas $26.1\%$, $3.7\%$, $2.3\%$ ($32.3\%$ in all) perfect ranking cases on average can only solved by the approach \techname, CoaCor and CODE-NN respectively. Such $32.3\%$ cases show the potential improvements on model combination, and this is why the model combination works well in our approach. 

To help understand the model combination, we also conducted an additional case study. In this case study, we analyze \techname with the existing state-of-the-art model (CoaCor, QN-RL$^\text{MRR}$ + CODE-NN).

\paragraph{\textbf{Case study.}} Table~\ref{tab:examples_ocor} shows three examples that are ranked perfectly by \techname but not CoaCor. As shown, there are many overlaps between the input natural language description and the code in SQL (e.g., the word ``select'' and ``data'' in the first example, row 2,3; the word ``table'' and ``time'' in the second example). In these examples, the information of overlap scores is important, where a human can utilize this information and retrieved the target code easily, and \techname catches such information successfully. Existing approaches like CoaCor do not utilize the information of overlap scores properly, where the CoaCor approach directly uses the token-level embedding for the neural network and replaces identifier names with numbered placeholder tokens (e.g., the SQL code in the second example is turned to ``select col0 (col1) from tab1'' in Table~\ref{tab:examples_ocor}). Thus, \techname has a good performance on these examples, while the CoaCor does not work well, which is the strength of \techname.

\begin{table}
  \begin{tabular}{lp{6cm}}
    \toprule
    Type&Example\\
    \midrule
    \midrule
    Description & how to use max and top in sql query in oracle?\\
    \midrule
    SQL Code & select	id,	item,	quantity,	date	from	(select	id,	item,	quantity,	date	from	your\_table	order	by	quantity	desc,	date	desc)	where	rownum	=	1;\\
    \midrule
    \midrule
    Description & find 1 level deep hierarchical relationship between columns of a table for one of the top level values\\
    \midrule
    SQL Code & select	t2.cat\_id,	t2.subcat\_id,	t2.name	from	test	t1	join	test	t2	on	t1.cat\_id	=	t2.cat\_id	where	t1.subcat\_id	=	42	and	t2.subcat\_id	<>	42	;\\
    \bottomrule
\end{tabular}
  \caption{The examples that ranked perfectly by CoaCor but not \techname.}
    \label{tab:examples_coacor}
\end{table}

To understand the weakness of \techname, we also conducted another case study on examples where \techname does not work well compared with the CoaCor. These examples are presented in Table~\ref{tab:examples_coacor}. As shown, in these examples there are few overlaps between the input natural language description and the code in SQL. In such a situation, \techname can hardly measure the overlap scores in the matrix, which makes it difficult to utilize the key information of overlap scores in \techname. However, CoaCor, which combines the annotation generation and code retrieval together, replaces identifier names with numbered placeholders, and extracts the high-level information of these situations. This is the reason why CoaCor has a good performance on these examples. The cases show the weakness of \techname, which does not have a good performance when the overlap scores are hard to measure. It is probably a good way to combine different approaches together and utilize the strength of each approach. Thus, we use the model combination to combine the strengths of different approaches. 

\begin{tcolorbox}
Answer to \textbf{RQ3}: The three techniques complements each other, allowing the model combinations to produce better results.
\end{tcolorbox}

\section{Threats to Validity}
\paragraph{\textbf{Threats to internal validity.}} A threat to internal validity is the potential faults in the implementation of our experiments.
To reduce this threat, for the performance of original models, we directly use their reported performances~\cite{Yao_2019}, and, for the performance of combined models, we directly used their published code~\cite{Yao_2019}. Furthermore, the implementation of our model was based on a published model~\cite{DBLP:journals/corr/abs-1911-09983} to avoid faults in re-implementation. 
\paragraph{\textbf{Threats to external validity.}} Our model was evaluated on the StaQC and the C\# benchmarks, which are widely used in previous code retrieval approaches~\cite{Yao_2019, iyer2016summarizing}. In these two benchmarks, all of the programs are collected from Stack Overflow, which gives a threat to external validity. However, we find that the overlap relations we used also widely exist in datasets collected from other sources such as GitHub~\cite{github}. For example, in CodeSearchChallenge Corpus~\cite{husain2019codesearchnet} which is another code retrieval benchmark collected from GitHub repositories, 98.6\%, 95.4\%, 94.18\% and 96.8\% of the instances have at least one overlap for Python, Java, JavaScript, and Go, respectively, while only 89.12\% in the StaQC dataset used in our experiment. Such widely existed relations show the potential value of our approach when applied to the GitHub benchmarks.  Meanwhile, since our approach was only tested on SQL and Java programs collected from Stack Overflow, further studies are also needed to apply our model to other programming languages collected from GitHub.

\section{Related work}
\paragraph{\textbf{Code Retrieval.}} Code retrieval in software development helps developers reuse the relevance code snippets among a large scale open-source projects. Early studies mostly focus on applying the information retrieval methods to code retrieval task~\cite{f5cf40c597ba4b0ead6630fd4cbd2607, Haiduc2013Automatic, Hill2014NL, gu2018deep, Keivanloo:2014:SWC:2568225.2568292, 7081874, inproceedingss}. With the development of deep learning, more and more works try to use neural networks to code retrieval~\cite{iyer2016summarizing,gu2018deep,Yao_2019}. 
\citeauthor{gu2018deep}~\cite{gu2018deep} first proposed an LSTM-based RNN for code retrieval, where they encode the input natural language description and code into a vector space and measure the cosine similarity between them. Based on a code annotation work proposed by \citeauthor{iyer2016summarizing}~\cite{iyer2016summarizing}, where they use a sequence-to-sequence model to generate the specific annotation by a given code, \citeauthor{Yao_2019}~\cite{Yao_2019} proposed CoaCor for code retrieval, where they combine the code annotation approach of \citeauthor{iyer2016summarizing} and the code retrieval approach of \citeauthor{gu2018deep} together by a reinforcement learning framework. Different from these approaches, we focus on unitizing the overlap scores between the natural language description and code. Based on this, we proposed a novel neural architecture for code retrieval. 

\paragraph{\textbf{Overlap Information}} Many works focus on using neural networks combined with overlap information in sentence pairs matching ~\cite{hu2015convolutional,Jian2016Convolutional,Hua2016Pairwise}. \citeauthor{hu2015convolutional}~\cite{hu2015convolutional} first proposed to use a neural network. They adopted a stack of convolution layers to infer the relation between the question and the given answer. \citeauthor{QiuH15}~\cite{QiuH15} introduced a transformation layer to use the interaction between the question and the answer. They tried to utilize hidden units to extract the overlaps based on hidden states. \citeauthor{Jian2016Convolutional}~\cite{Jian2016Convolutional} proposed a kind of overlap features and combined it with a convolutional network. Such overlap features compute the similarity between two words via whether they are the same. It cannot utilize the overlap between words (e.g., the overlap between ``joint\_table\_a'' and ``table''), which is important in encoding identifiers in code. The atomic value of these overlap features cannot represent the relation between identifiers and corresponding words. Thus, we design the overlap matrix based on longest common sub-string to measure the degree of the overlap. We also adopt some special neural components for this representation.
\section{Conclusion and Future work}
In this paper, we proposed a novel overlap-aware neural architecture (\techname) for code retrieval. Our approach makes use of the overlap score between the natural language description and the code by using the overlap matrix and the character embedding. 

We evaluated our approach on several datasets. The experimental results show that \techname achieves significant improvement compared with existing state-of-the-art approaches. 
The further evaluation shows that 
each component in our approach is important.
\paragraph{\textbf{Future work.}}
Our approach is built mainly on the basis of overlap scores between two inputs, especially for the natural language description and code. The experimental results show that the overlap score can boost the performance of the model. It is interesting to study as the further work to try more metrics to measure the degree of the overlaps between two strings (e.g., Edit Distance and Longest Common Sub-string). Our experiment also shows the potentiality of the ensemble (namely model combination), which may also be an effective way to use such technique to combine different metrics to improve the performance.
Furthermore, \techname, can be directly applied to other programming languages (e.g., Java, Python and C++). Our model is designed for the general code retrieval task and some specific features may be added to it by gating. It is also interesting to study as the further work.

\section*{ACKNOWLEDGMENTS}
This work is sponsored by the National Key Research and Development Program of China under Grant No. 2017YFB1001803, and National Natural Science Foundation of China under Grant Nos. 61672045 and 61922003.
\bibliographystyle{ACM-Reference-Format}
\bibliography{test}

\end{document}